# Large language models have learned to use language

Commentary on Futrell & Mahowald's How Linguistics Learned to Stop Worrying and Love the Language Models (*BBS*, Forthcoming)

Gary Lupyan
University of Wisconsin-Madison
1202 W. Johnson St. Madison, WI 53706
lupyan@wisc.edu
http://sapir.psych.wisc.edu

## Abstract

Acknowledging that large language models have learned to use language can open doors to breakthrough language science. Achieving these breakthroughs may require abandoning some long-held ideas about how language knowledge is evaluated and reckoning with the difficult fact that we have entered a post-Turing test era.

It is instructive to compare Futrell and Mahowald's well-reasoned and empirically grounded target article with counterpoints from skeptics demanding that LLMs conform to theories of generative linguistics to be judged as good models of language (Bolhuis et al., 2024; Murphy et al., 2025), or who insist that LLMs are "stochastic parrots" that "only manipulate the form of language, with neither understanding nor communicative intent" (Bender & Hanna, 2025). Gauging the usefulness of LLMs as models of language based on how well they adhere to a particular theory of language is not only self-serving and unscientific, but risks further isolating many linguists from the broader language sciences. And reducing LLMs to "stochastic parrots" is an insult to people's intelligence. Are we to believe that the 800M weekly users of ChatGPT are credulous dolts unable to distinguish competent language use from mere parroting? Or is it more likely that a decades-long scientific program of learning language through prediction (Elman, 1990) was on to something; that learning language from data is not only possible, but actually works?

By actually works I mean that a general-purpose neural network endowed with a simple learning rule to lower prediction error, an architectural capacity to track long-distance associations, and a corpus of language use, can learn language. Some continue to deny the premise, arguing that the failure of LLMs to "Draw [a] tree structure, in line with Minimalist syntax, for the sentence 'The doctor the nurse the hospital had hired met John'" means that LLMs lack "fundamental principles of linguistic structure" (Murphy et al., 2025). This argument is absurd. My 5-year-old is a highly competent language user who cannot draw a parse tree or tell you what makes a sentence ungrammatical. What counts is use! If a language model can use language proficiently enough to pass the Turing test (Jones & Bergen, 2025) while failing to conform to minimalist syntax, so much the worse for minimalist syntax (Hu et al., 2024).

The ability to judge sentences like "Colorless green ideas sleep furiously" to be grammatical despite being nonsensical, has played an outsized role in generative linguistics. These metalinguistic judgments are often used to test competing theories (Schütze, 2016), and have often been used to evaluate LLMs (e.g., Dentella et al., 2023). That children engage in sophisticated language use while lacking the ability to make such explicit judgments is *prima facie* evidence that being a competent language user does not require being able to dissociate conceptual content from structural configurations. It is also true, however, that
many adults without linguistic training can do this. But so can LLMs. As we previously showed, ChatGPT-4 is better than people at judging normative grammaticality of sentences (Hu et al., 2024). In recent informal tests, we also found that GPT5 can *generate* meaningless grammatical sentences as well as adults. Figure 1 shows meaningfulness ratings of sentences generated by people and GPT5 (t<1), as well as their attempts to guess whether each sentence was generated by a person or an LLM (t<1).

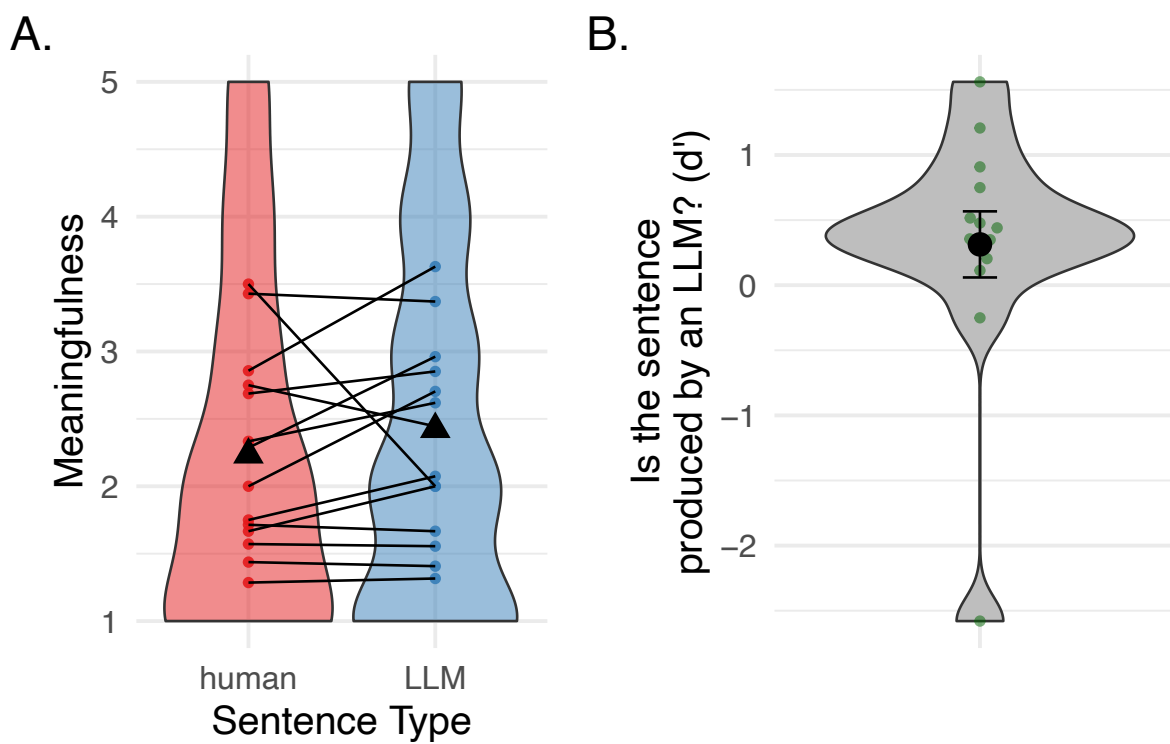

*Figure 1*: A. Results from an informal experiment in which participants were tasked with generating maximally meaningless grammatical sentences. Each participant then judged the meaningfulness of the sentences generated by other participants and ChatGPT5 and whether each sentence was generated by a person or an LLM. (A) Mean meaningfulness ratings with lines joining each participant's mean ratings. (B) Accuracy in judging whether each sentence was generated by a person or LLM. An example of similarly meaningless LLM and human sentences: "Let velvet measure tomorrow with raisins"; "Burpy trees monkey in mountain poodle".

An intriguing question is when the ability to distinguish meaningfulness from grammaticality emerges in LLMs and whether it has any causal bearing on proficient language use. Although grammaticality can be read out of base models (LLMs trained only through self-supervision) using surprisal, base models appear incapable of making explicit grammaticality judgments (Hu et al., 2024). In our preliminary tests, we find that base models can generate implausible sentences (e.g., A cat barks at the dog), but struggle to generate ungrammatical or truly meaningless ones. The ability to explicitly uncouple meaningfulness from grammaticality appears to emerge with fine-tuning on even small datasets (e.g., Databricks-dolly-15k 2023) which do not contain any examples of grammaticality judgments. What makes this intriguing is that despite their enormous latent knowledge of both language and the world, base models make lousy conversationalists (the public couldn't care less about GPT-3 but was blown away by its fine-tuned version released as ChatGPT). Understanding whether the ability to make explicit grammatical judgments is causally related to being able to *use* language would be a fascinating development.

It seems that over the decades, some linguists have lost the plot (see Knight, 2016 for an engaging intellectual history). The goal of a language user is not to generate parse trees that conform to minimalist syntax. Rather, it is to *do* things with language. This includes inferring people's mental states from what they say, communicating goals to enable collaboration, and convincing them to act in your interests. To the extent that such language use *requires* learning specific aspects of language structure, we should find these by appropriately interrogating LLMs. For example, it seems clear that representing the hierarchical structure—in some form—is critical to parsing language, and converging evidence appears to show that in learning language LLMs learn hierarchical structure (e.g., Goldstein et al., 2025; Liu et al., 2025). LLMs provide an unprecedented opportunity to understand the aspects of linguistic structure are critical for language use, even when learned by a system with an architecture radically different from ours.

In his 1990 short story *Bears Discover Fire*, Terry Bisson imagines a world just like ours, but in which, inexplicably, bears have discovered fire: "They don't hibernate anymore … They make a fire and keep it going all winter" (Bisson, 1990). One reaction to such a development would be to insist that bears are just not the kind of animal that can control fire; that it's a trick; that their fire must not be *real* fire. Another would be to wonder if perhaps we were mistaken in our understanding of what bears are capable of, or in our understanding of what it takes to control fire. We have now entered a post Turing-test era where the torch of language is no longer unique to our species. What we make of this new reality is up to us.

**Acknowledgment**

The author would like to thank Zach Studdiford for his help on comparing Gemma base and instruction-tuned language models.

**Competing Interests**

None

**Funding**

This work was partially supported by NSF-PAC 2020969 to G.L.

**References**

Bender, E. M., & Hanna, A. (2025). *The AI Con: How to Fight Big Tech's Hype and Create the Future We Want*. Harper.

Bisson, T. (1990, 2014). Bears Discover Fire. *Lightspeed Magazine*, *44*. https://www.lightspeedmagazine.com/fiction/bears-discover-fire/

Bolhuis, J. J., Crain, S., Fong, S., & Moro, A. (2024). Three reasons why AI doesn't model human language. *Nature*, *627*(8004), 489–489. https://doi.org/10.1038/d41586-024-00824-z

*Databricks/databricks-dolly-15k · Datasets at Hugging Face*. (2023, October 1). https://huggingface.co/datasets/databricks/databricks-dolly-15k

Dentella, V., Günther, F., & Leivada, E. (2023). Systematic testing of three Language Models reveals low language accuracy, absence of response stability, and a yes-response bias. *Proceedings of the National Academy of Sciences*, *120*(51), e2309583120. https://doi.org/10.1073/pnas.2309583120

Elman, J. L. (1990). Finding Structure in Time. *Cognitive Science*, *14*(2), 179–211.

Goldstein, A., Ham, E., Schain, M., Nastase, S. A., Aubrey, B., Zada, Z., Grinstein-Dabush, A., Gazula, H., Feder, A., Doyle, W., Devore, S., Dugan, P., Friedman, D., Brenner, M., Hassidim, A., Matias, Y., Devinsky, O., Siegelman, N., Flinker, A., … Hasson, U. (2025). Temporal structure of natural language processing in the human brain corresponds to layered hierarchy of large language models. *Nature Communications*, *16*(1), 10529. https://doi.org/10.1038/s41467-025-65518-0

Hu, J., Mahowald, K., Lupyan, G., Ivanova, A., & Levy, R. (2024). Language models align with human judgments on key grammatical constructions. *Proceedings of the National Academy of Sciences*, *121*(36), e2400917121. https://doi.org/10.1073/pnas.2400917121

Jones, C. R., & Bergen, B. K. (2025). *Large Language Models Pass the Turing Test* (No. arXiv:2503.23674). arXiv. https://doi.org/10.48550/arXiv.2503.23674

Knight, C. (2016). *Decoding Chomsky: Science and Revolutionary Politics*. Yale University Press.

Liu, W., Xiang, M., & Ding, N. (2025). Active use of latent tree-structured sentence representation in humans and large language models. *Nature Human Behaviour*, 1–14. https://doi.org/10.1038/s41562-025-02297-0

Murphy, E., Leivada, E., Dentella, V., Montero, R., Günther, F., & Marcus, G. (2025). Fundamental Principles of Linguistic Structure Are Not Represented by ChatGPT. *Biolinguistics*, *19*, 1–55. https://doi.org/10.5964/bioling.19021

Schütze, C. T. (2016). *The empirical base of linguistics: Grammaticality judgments and linguistic methodology*. Language Science Press. http://www.oapen.org/search?identifier=603356